%% file: acl_latex.tex
\documentclass[11pt]{article}

\usepackage[final]{acl}

\usepackage{times}
\usepackage{latexsym}

\usepackage[T1]{fontenc}

\usepackage[utf8]{inputenc}

\usepackage{microtype}

\usepackage{inconsolata}

\usepackage{graphicx}
\usepackage{microtype}

\usepackage{mathtools}
\usepackage{amsthm}
\usepackage{makecell}
\usepackage{amsmath}
\usepackage{graphicx}
\usepackage{multirow}
\usepackage{subfigure}
\usepackage{tcolorbox}
\usepackage{amssymb}
\usepackage{fontawesome5}
\usepackage{marvosym}
\usepackage{lineno}
\usepackage{listings}
\usepackage{colortbl}
\usepackage[table]{xcolor}
\usepackage{booktabs}   
\usepackage{tabularx}   
\usepackage{pifont}     
\usepackage{enumitem}
\definecolor{ForestGreen}{RGB}{84, 209, 120}
\definecolor{badred}{RGB}{180, 0, 0}
\definecolor{goodgreen}{RGB}{0, 100, 80}
\newcommand{\cmark}{\textcolor{goodgreen}{\ding{51}}} 
\newcommand{\xmark}{\textcolor{badred}{\ding{55}}}   

\newcommand{\datasetname}{\textbf{UnivEARTH }}
\newcommand{\datasetnamenospace}{\textbf{UnivEARTH}} 
\newcommand{\syndatasetname}{\textbf{OpenEARTH }}

\title{Towards LLM Agents for Earth Observation}

\author{
  \textbf{Chia-Hsiang Kao\textsuperscript{1}},
  \textbf{Wenting Zhao\textsuperscript{1}},
  \textbf{Cheryl Lam\textsuperscript{1}},
\\
  \textbf{Aarush Umap\textsuperscript{1}},
  \textbf{Shreelekha Revankar\textsuperscript{1}},
  \textbf{Samuel Speas\textsuperscript{1}},
\\
  \textbf{Snehal Bhagat\textsuperscript{1}},
  \textbf{Rajeev Datta\textsuperscript{1}},
  \textbf{Cheng Perng Phoo\textsuperscript{1}},
\\
  \textbf{Utkarsh Mall\textsuperscript{3}},
  \textbf{Carl Vondrick\textsuperscript{2}},
  \textbf{Kavita Bala\textsuperscript{1}},
  \textbf{Bharath Hariharan\textsuperscript{1}}
\\
\\
  \textsuperscript{1}Cornell University,
  \textsuperscript{2}Columbia University,
  \textsuperscript{3}MBZUAI
\\
  \small{
    \textbf{Correspondence:} \href{mailto:ck696@cornell.edu}{ck696@cornell.edu}
  }
}

\begin{document}
\maketitle

\begin{figure*}[h]
  \centering
  \includegraphics[width=0.95\linewidth]{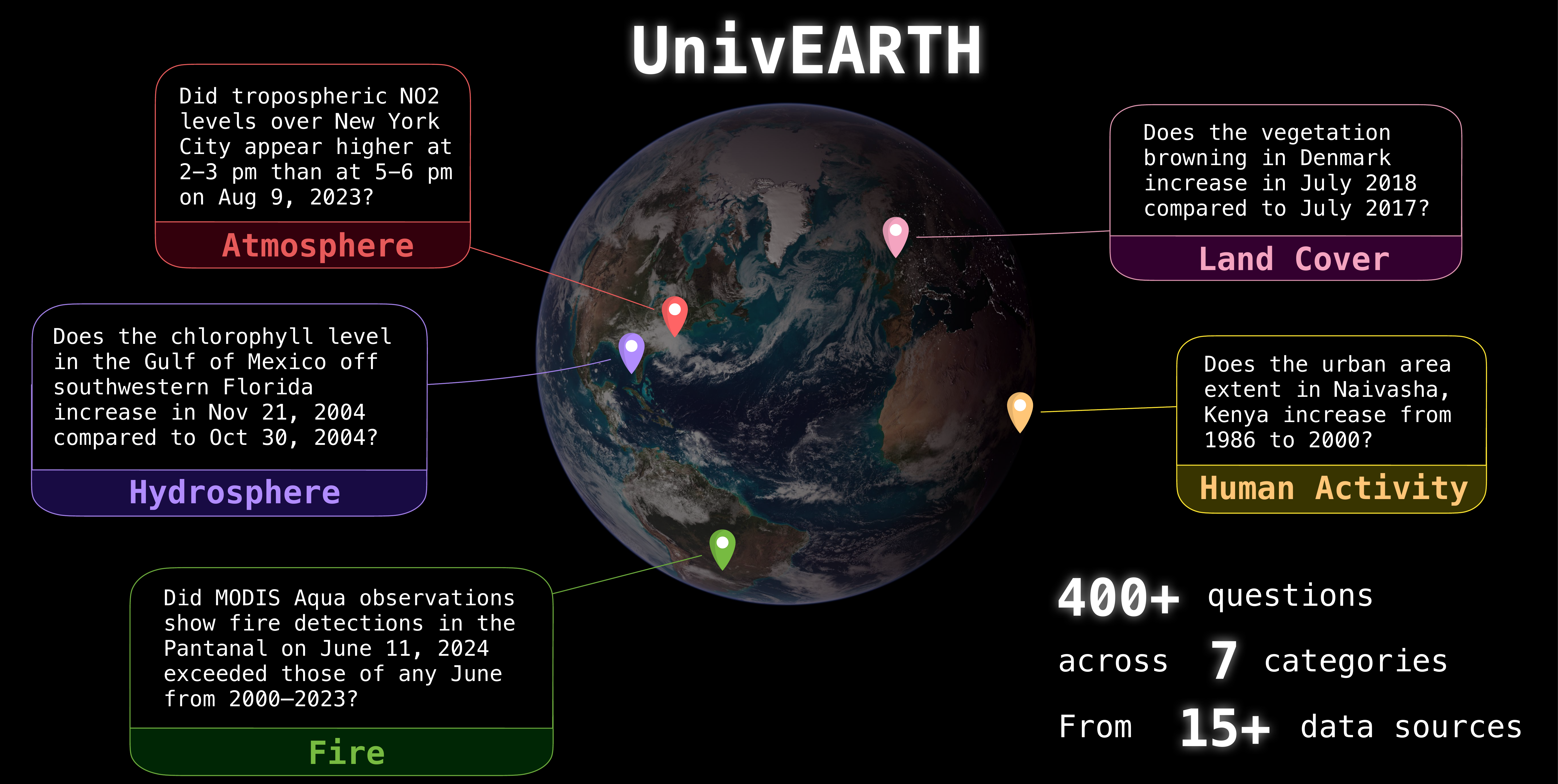}
  \vspace{-1mm}
  \caption{We propose \datasetname for benchmarking AI agents in Earth Observation.}
  \label{fig:teaser}
\end{figure*}

\begin{abstract}
\input{sec/main_0_abstract}
\end{abstract}

\input{sec/main_1_introduction}
\input{sec/main_2_literature_review}
\input{sec/main_3_dataset_curation}
\input{sec/main_4_method}
\input{sec/main_5_results}
\input{sec/main_6_conclusion}

\section*{Limitations}

We acknowledge several limitations of \datasetname.
First, the current benchmark does not include unanswerable questions~\citep{rajpurkar2018know, kim20222}, where the ground-truth answer is ``inconclusive.'' 
Second, \syndatasetname is limited to a yes–no question format; future versions could be extended to include multiple-choice scenarios. 
Finally, most questions are mined from NASA websites; future iterations could broaden coverage by sourcing questions from scientific papers.




\bibliography{custom}


\newpage
\appendix

\label{sec:appendix}
\input{sec/main_7_appendice}
\input{sec/main_7_annotation}

\end{document}

%% file: sec/main_0_abstract.tex
Earth Observation (EO) provides critical planetary data for environmental monitoring, disaster management, climate science, and other scientific domains.
In this work we ask: \textit{Are AI systems ready for reliable Earth Observation?}
To answer this, we introduce \datasetnamenospace, a coding benchmark of 408 yes/no questions from NASA Earth Observatory articles across 7 various topics and over 15 satellite instruments and sources. 
Using Google Earth Engine API as a tool in a zero-shot setup, LLM agents achieve an accuracy of 40.0\% where the code fails to run over 44\% of the time. 
To better understand LLM agent behavior, we also analyze the impact of using the JavaScript API versus Python and the effect of providing documentation. Furthermore, we find that using a reflexion~\cite{shinn2023reflexion} framework significantly reduces errors: Claude-4.5-Sonnet, Gemini-2.5-Pro, and GPT-5 accuracies rise to around 60\%. However, these results remain only marginally above random chance.
Taken together, our findings identify significant challenges to be solved before AI agents can automate earth observation, and suggest paths forward. 
Our dataset is publicly available.\footnote{\url{https://iandrover.github.io/2025_univearth/}}

%% file: sec/main_1_introduction.tex
\section{Introduction}

In a range of academic disciplines from plant science to anthropology, scientists routinely find the need to analyze planetary data: data about land use, earth surface reflectance, chlorophyll content, and so on.
This planetary data is collated and processed from a multitude of "Earth Observation" satellites, and the scientific process involves carefully choosing the right sensor, product, location, and time.
Automating this analysis would thus significantly accelerate the scientific process.

Our goal in this paper is to explore AI systems that can automate the task of earth observation in these scientific workflows and thus accelerate the scientific process.
While specialized automatic systems for specific earth observation tasks have been deployed for years ~\citep{watch2002global, giglio2016collection, wu2018global}, they lack the flexibility needed for general-purpose, customized queries.

Recent efforts to combine agentic frameworks with remote sensing have largely relied on pre-trained models and predefined toolsets~\cite{liu2024change, xu2024rs, shabbir2025thinkgeo, feng2025earth}. While these approaches simplify benchmarking, they often fall short of capturing "in-the-wild" complexities. Real-world analysis introduces significant challenges, including the ambiguity of diverse data sources (e.g., Terra vs. Aqua, Landsat-8 vs. 9), sensor-specific spectral bands, and constraints such as revisit rates and cloud cover. By relying on curated environments, prior studies often fail to test applicability in more real-world, dynamic settings.
We therefore ask:
\textit{\textbf{Are AI systems ready for reliable Earth Observation?}}

With these desiderata in mind, we begin by introducing \datasetname: a question-answering (QA) benchmark designed to evaluate LLMs for earth observation.
There are two challenges in building such a benchmark: (1) we need to know the kind of questions that one might ask about earth observation data and the corresponding answers, and (2) we need to ensure that the evidence or data needed to support the answer exists and is available.
Unlike existing benchmarks, such data of questions, answers, and supporting evidence is not freely available.
We address this challenge by leveraging a unique public resource: articles from the \emph{NASA Earth Observatory}~\citep{nasa}.
Each article walks through conclusions derived from observations from satellite imagery.
We rigorously curate question-answer pairs from these articles by: (1) leveraging LLMs alongside manually curated QA examples, (2) verifying question answerability through Google Earth Engine (GEE), and (3) careful independent review of each example. The resulting dataset, \datasetnamenospace, comprises 408 high-quality yes/no questions spanning 7 diverse topics and 15 main sources.



We benchmark several state-of-the-art models, including Gemini~\cite{team2023gemini}, Claude, GPT, DeepSeek~\cite{guo2025deepseek}, and Kimi~\cite{team2025kimi}. In our setup, simply answering the question is not enough: the models must ground their answers in evidence. We ask models to generate executable code using the Google Earth Engine~\citep{gorelick2017google} API to derive quantitative results that support their answers.
Our evaluation reveals significant limitations in current models. We find that agents fail to produce executable code a significant portion of the time, resulting in a best-case baseline accuracy of only 40.0\%.
To better understand these failures, we analyze performance across various dimensions. We find that: (1) the choice of API language (Python vs. JavaScript) slightly impacts model performance, while logic errors persist across both; (2) providing specific documentation on sensor information (e.g., Collection IDs) drastically reduces syntax errors and hallucinations; and (3) implementing a \textit{reflexion} mechanism~\cite{shinn2023reflexion}, allowing agents to debug based on error logs, significantly recovers performance, boosting accuracy to over 60\%.

In sum, our work makes three key contributions:
\begin{itemize}
\item We curate \datasetnamenospace, a novel evaluation benchmark for Earth Observation derived from authoritative NASA sources with verified answers and executable ground truth.
\item We benchmark state-of-the-art LLMs, revealing that while they possess general knowledge, they struggle with the specific syntax and data retrieval logic required for EO analysis.
\item We demonstrate that advanced agentic strategies, specifically the injection of precise documentation and the use of error-based reflexion, can bridge the gap between generation failures and reliable scientific analysis.
\end{itemize}

%% file: sec/main_2_literature_review.tex
\section{Related Work}

\textbf{LLMs for Scientific Applications.}
Scientific question answering has garnered significant attention, demonstrated by the development of benchmarks across various domains. General scientific QA benchmarks assess reasoning across multiple scientific disciplines~\citep{saikh2022scienceqa, hendrycks2020measuring, wang2023scibench, liang2024scemqa, feng2024sciknoweval, wang2024t}, while specialized benchmarks focus on specific areas such as medicine and biology~\citep{he2020pathvqa, li2024mmsci}, chemistry and material science~\citep{jablonka2024leveraging,alampara2024probing, chen2025unveiling}, and remote sensing~\citep{wang2024earthvqa, danish2024geobench, li2024show}.

Many of these prior benchmarks rely on models' internal knowledge, which may not be sufficiently rigorous in a scientific domain. In contrast, \datasetname demands grounding answers in evidence derived from satellite imagery and products, requiring more interpretable and explicit reasoning.
In this vein, our work is similar to prior work on leveraging existing tools or databases~\citep{m2024augmenting, fossi2024swiftdossier, campbell2025mdcrow, laurent2024lab}, but requires models to navigate a much larger repertoire of data sources (here, sensors and products).
These capabilities are a necessary first step if one seeks to automate discovery in the earth sciences, as prior work has sought to do for chemistry~\citep{zheng2025integrating, chen2025auto},  biology~\citep{swanson2024virtual}, or material science~\citep{strieth2024delocalized}.

\vspace{1mm}
\noindent \textbf{Code Generation and Tool-Using AI.}
Outside of scientific applications, several benchmarks evaluate code generation capabilities, including SWE-bench~\citep{jimenez2023swe}, SWT-Bench~\citep{mundler2024swt}, LiveCodeBench~\citep{jain2024livecodebench}, and SWE-bench Multimodal~\citep{yang2024swe}. These benchmarks primarily focus on general software engineering tasks rather than domain-specific scientific applications.
In the context of data analysis, text-to-SQL benchmarks like Spider~\citep{yu2018spider}, SEDE~\citep{hazoom2021text}, BIRD~\citep{li2023can}, and Spider 2.0~\citep{lei2024spider} evaluate models' ability to translate natural language questions into database queries. 

\vspace{1mm}
\noindent \textbf{Earth Observation With AI Agents.}
Recently, there has been a surge in research applying agentic frameworks to remote sensing and satellite imagery analysis. For instance, Change-Agent~\cite{liu2024change} and RS-Agent~\cite{xu2024rs} integrate LLMs with pre-trained vision models to address tasks such as change detection, object detection, and scene classification. Similarly, ThinkGeo~\cite{shabbir2025thinkgeo} and EarthAgent~\cite{feng2025earth} utilize predefined functions and tools to execute tasks and benchmark their performance. However, these approaches often rely on highly curated environments that fail to capture 'in-the-wild' complexities. Real-world scenarios introduce significant ambiguities, including diverse data sources (e.g., Terra/Aqua, Landsat-8/9, or Sentinel-2), sensor-specific spectral bands, and availability constraints such as revisit rates and cloud cover. Previous work simplifies the complexities of real-world problems and thus limits its applicability in more dynamic, real-world settings. \datasetname extends this paradigm to a realistic Earth Observation domain, designed to handle the complexities of accessing and analyzing satellite data.

%% file: sec/main_3_dataset_curation.tex
\section{Benchmark Construction}

\begin{figure}[t]
  \centering
  \includegraphics[width=0.9\linewidth]{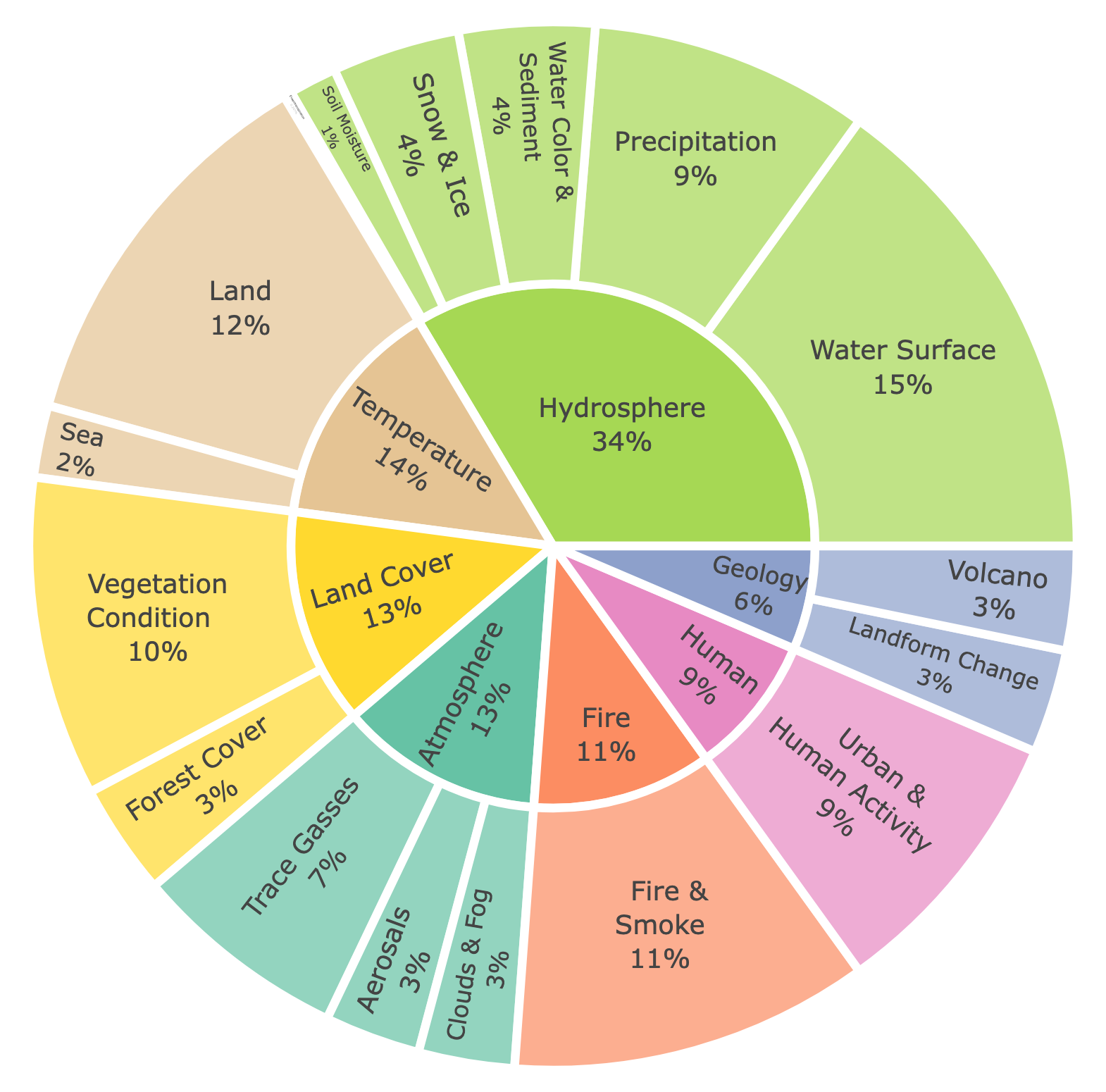}
  \vspace{-1mm}
  \caption{Hierarchical distribution of Earth science topics and sub-categories in \datasetname. The inner ring represents the parent domain, while the outer ring details specific phenomena.}
  \label{fig:topic_dist}
\end{figure}

\input{tables/sec3_result_modality_count}

\subsection{Data Source}

EO science relies heavily on the analysis of remotely sensed data to investigate changes and phenomena. This characteristic makes it particularly suitable for automation through AI agents that can process and analyze large volumes of imagery data. However, benchmarking such AI capabilities requires high-quality question-answer pairs that are both scientifically sound and verifiable through available data sources.
To develop our benchmark, we identified NASA's Earth Observatory website~\citep{nasa} as an authoritative primary source. Since its inception in May 1998, this platform has published articles covering diverse topics including air quality, climate change, human impact monitoring, and natural events. These articles, authored by NASA Earth Observatory's science writers, provide reliable scientific reporting based on imagery analysis and research findings.

\subsection{Data Collection and Validation}

\input{tables/sec3_dataset_example}

The curation pipeline comprises of three stages: annotation, verification, and review. 

\vspace{1mm}
\noindent \textbf{Annotation.} 
We focus on NASA Earth Observatory website articles with the cutoff time of September 30, 2025. Three annotators generate candidate yes/no question-answer pairs with supporting sentences. We chose the yes/no question format because assessing the correctness of free-form answers is challenging and less objective in scientific domains. During this process, we disregard articles about sensor specifications, general introductions, non-satellite imagery, transient observations (e.g., wind speed, tides), or statements without clear sources. Table~\ref{tab:dataset_example} shows some example questions and supporting sentences.

\vspace{1mm}
\noindent \textbf{Verification.} We examined whether questions derived from NASA articles can be answered using the data available in Google Earth Engine (GEE)~\citep{gorelick2017google}. Background details on GEE are presented in Appendix-\ref{app:gee_intro}. 
This verification step was critical for ensuring the questions in the benchmark are approachable with existing data catalog hosted in GEE.
This was necessary because some articles describe phenomena using sensors or products not available in GEE, making it impossible for agents to answer them.
For example, articles (e.g., on~\href{https://earthobservatory.nasa.gov/images/87182/sulfur-dioxide-down-over-the-united-states}{December 2015},~\href{https://earthobservatory.nasa.gov/images/88153/satellite-finds-unreported-sources-of-sulfur-dioxide}{June 2016}, or~\href{https://earthobservatory.nasa.gov/images/90276/the-ups-and-downs-of-sulfur-dioxide-in-north-america}{May 2017}) about sulfur dioxide often reference the Ozone Monitoring Instrument (OMI) on the Aura satellite. While OMI has monitored sulfur dioxide and other air pollutants since 2004, its sensor data is not available on GEE. The alternative Sentinel-5P NRTI sulfur dioxide datasets only begin in 2018, creating a temporal constraint.

\vspace{1mm}
\noindent \textbf{Review.} 
Following verification, we recruited reviewers to evaluate the quality and clarity of the questions. These reviewers were asked to:
(Q1) Provide a yes/no answer to each question based on the text and image of the article;
(Q2) Assess whether the answer was supported by the text in the corresponding NASA article;
(Q3) Evaluate whether the answer was supported by imagery in the article; and
(Q4) Assess if location information needs verification through external sources. The fourth assessment point was included because some questions, particularly those manually edited or designed, required geographical review. In these cases, reviewers were permitted to use Google Maps to verify geo-locations. 
Please refer to Appendix.~\ref{app:evaluation_criteria} for the reviewer instruction document.

In the first version of the dataset, composed of 140 questions, four reviewers were recruited, with each reviewer evaluating half of the dataset.
The initial review showed inter-reviewer agreement rates of 90.1\%, 73.2\%, 78.9\%, and 81.7\% for Q1, Q2, Q3, and Q4, respectively. The agreement rate is computed as an exact match. Following this initial assessment, we iteratively revised ambiguous questions with each reviewer until we reached complete agreement on Q1. In the extended version of the dataset, composed of 268 questions, each question is reviewed by two reviewers, and the question-answer pairs are iteratively revised until reviewers and annotators reach consensus. 
Because the review process focused on verifying the correctness of answers with respect to the source NASA articles rather than domain-specific expertise, we recruited seven reviewers who were undergraduate or graduate students in computer science.

\subsection{Dataset Statistics}

Figure~\ref{fig:topic_dist} illustrates the distribution of topics within the dataset. The largest category is \textit{hydrosphere} (34\%), encompassing questions regarding surface water extent, precipitation, water color, snow, and moisture. Other major categories include temperature (14\%, primarily land surface temperature), land cover (13\%, such as vegetation and forest cover), and atmosphere (13\%, including trace gases, aerosols, and clouds). Additional domains cover fire, human activities (e.g., urban expansion), and geology. This broad range of tasks presents a holistic view of Earth Observation, marking a prominent distinction from previous works that primarily focus on object detection or scene classification.
Complementing this, Table~\ref{tab:source_dist} details the major data sources used in our dataset. The majority of observations are derived from Landsat and the MODIS system (onboard the Terra and Aqua satellites), and VIIRS.

\subsection{Relevance to Science and Real-World Impact}
\datasetname captures phenomena with significant real-world relevance and active scientific interest. For instance, 
our benchmark includes questions about red tides in the Gulf of Mexico derived from \href{https://earthobservatory.nasa.gov/images/5071/red-tide-off-florida}{a December 2004 article}, addressing phenomena linked to potential fish mortality and human respiratory irritation. 
Another question focuses on the number of lakes on the Tibetan Plateau based on \href{https://earthobservatory.nasa.gov/images/154011/a-proliferation-of-lakes-on-the-tibetan-plateau}{a March 2025 article}, directly connecting to recent research on accelerated lake formation in this critical region~\citep{li2022variation, lei2023unprecedented, zhou2024accelerating, zhou2024annual}. 
The benchmark also covers other scientifically relevant topics including 
chlorophyll concentration and climate patterns in the Pacific Ocean~\citep{wang2005ecosystem},
the trend of disappearing lakes in Siberia~\citep{smith2005disappearing}, 
lake surface albedo dynamics~\citep{argaman2012monitoring}, groundwater depletion in the Indus Basin~\citep{richey2015quantifying}, 
increasing global leaf area~\citep{chen2019china}, 
and global cropland expansion~\citep{potapov2022global}. 
Thus our benchmark provides a sampling of questions that scientists may want answered in the course of their research.

%% file: tables/sec3_result_modality_count.tex
\begin{table}[t]
\centering
\small
\caption{Distribution of the main data sources grouped by sensing modality and instrument type.}
\label{tab:datasources}
\vspace{-1.7mm}
\definecolor{headerblue}{rgb}{0.9, 0.95, 1.0}
\begin{tabular}{lr}
\toprule
\textbf{Category -- Instrument / Source} & \textbf{Count} \\
\midrule
\rowcolor{headerblue} \multicolumn{2}{l}{\textbf{\textit{Multispectral Optical Imaging}}} \\
Landsat Series (4--9) & 162 \\
MODIS System (Terra/Aqua) & 140 \\
VIIRS & 60 \\
SeaWiFS & 3 \\
Sentinel-2 & 2 \\
\midrule
\rowcolor{headerblue} \multicolumn{2}{l}{\textbf{\textit{Atmospheric Composition}}} \\
Sentinel-5P (TROPOMI) & 16 \\
TOMS & 4 \\
OMI-Aura & 4 \\
TEMPO & 2 \\
\midrule
\rowcolor{headerblue} \multicolumn{2}{l}{\textbf{\textit{Hydrology \& Cryosphere}}} \\
GRACE & 11 \\
GPM & 8 \\
CHIRPS & 7 \\
SMAP & 4 \\
\midrule
\rowcolor{headerblue} \multicolumn{2}{l}{\textbf{\textit{Weather Satellites \& Models}}} \\
GEOS Model & 17 \\
ERA5 Reanalysis & 3 \\
\midrule
\textbf{Total} & \textbf{443} \\
\bottomrule
\end{tabular}
\label{tab:source_dist}
\end{table}

%% file: tables/sec3_dataset_example.tex
\begin{table*}[htbp]
\centering
\caption{Examples of \datasetname.}
\vspace{-1.7mm}
\small
\label{table:category_examples}
\begin{tabular}{p {2.1cm} p{4.8cm} p{7.5cm}} 
\toprule
Topic & Example & Supporting Sentences \\
\midrule
\textbf{\textit{Atmosphere}}
& Did nitrogen oxide concentrations in the Northern Hemisphere increase from 2019 to 2020?
& The annual growth rate for 2020 was the highest scientists had recorded since systematic annual methane measurements began in 1983—an increase of 15 parts per billion, which was exceeded again in 2021.
\\
\midrule
\textbf{\textit{Land Cover}}
& Does forest cover decrease in Argentina's Salta Province from December 2000 to December 2019?
& The images above show deforestation over a span of two decades around the Salta Province of northern Argentina. The image from December 18, 2000, shows a mix of cleared land and greener areas. The image from December 24, 2019, shows much of the forest replaced by large fields.
\\
\midrule
\textbf{\textit{Hydrosphere}}
& Does Lake Erie have more ice coverage compared to the other Great Lakes in February 14, 2018 afternoon?
& On the same date last year, total ice cover was 9.7 percent. Lake Erie was the iciest of the five lakes, with 93.3 percent iced over.
\\
\bottomrule
\vspace{-1.5em}
\label{tab:dataset_example}
\end{tabular}
\end{table*}

%% file: sec/main_4_method.tex
\section{Benchmarking SoTA Agents with \datasetname}
In this section, we evaluate state-of-the-art LLM agents on our benchmark dataset.

\paragraph{Experimental Setup.}
We adopt a zero-shot setup for evaluation. Given a query, the model is instructed to first reason about the plan for the code and then generate a Google Earth Engine (GEE) Python script. This script is executed, and the resulting output, along with the original question and code, is processed by Gemini-2.5-Flash to derive the final output.

To address the high frequency of code generation failures (discussed subsequently), we employ a specific error taxonomy during the evaluation. We prompt Gemini-2.5-Flash to classify failures into one of the following categories:
\begin{itemize}[itemsep=0pt]
\item Empty Collection (C1): No images found for the specified date or location in one or both comparison points.
\item No Valid Pixels (C2): Images exist, but no valid pixels remain after preprocessing, such as cloud masking.
\item Calculation Failure (C3): The result is None, NaN, or otherwise invalid\footnote{For example, Claude-4.5-Sonnet produced a physically implausible value of 1723.29$^\circ$C by neglecting to apply the 0.01 scale factor inherent to the \texttt{NOAA/CDR/OISST/V2\_1} dataset.}.
\item Code/Syntax Error (D): Failures caused by incorrect image/band names, syntax errors, invalid methods, or memory issues.
\end{itemize}
This categorization allows for a detailed diagnosis of model failure modes in real-world scenarios.

Our primary metric is correctness, defined as the fraction of questions where the generated answer matches the ground truth. We also measure the wrong answer rate alongside the distribution of error types.
We benchmark a suite of LLM agents, including Gemini-2.5-Pro, Gemini-2.5-Flash~\cite{team2023gemini}, Claude-4.5-Sonnet, Claude-4.5-Haiku, GPT-5, DeepSeek-Reasoner~\cite{deepseekai2025deepseekr1incentivizingreasoningcapability}, DeepSeek-Chat, and Kimi-k2~\cite{team2025kimi}.

\input{tables/bl_gee}

\subsection{Answering Questions With Google Earth Engine}
\label{subsec:bl_gee}

\input{tables/collection_hallucination}

Table~\ref{table:gee_bl} presents the average performance across the dataset. The highest overall accuracy achieved was only~40.0\% by Claude-4.5-Sonnet, closely followed by Gemini-2.5-Pro (36.9\%) and GPT-5 (33.1\%). DeepSeek-Reasoner and DeepSeek-Chat exhibit comparable levels of correctness, whereas Kimi-k2, Claude-4.5-Haiku, and Gemini-2.5-Flash perform the poorest, with accuracy below 20\%.

A deeper investigation into failure modes reveals a persistently high rate of syntax and method errors (Category D). Even the strongest model, Claude-4.5-Sonnet, suffered a 21.4\% error rate in this category, while all others exceeded 30\%. This indicates that current models lack proficiency in utilizing the GEE API. We show some examples of hallucinated collection ID or names in Table~\ref{tab:id_hallucinations}.

Regarding specific execution failures, the most common issue is that there is no available image for the queried dataset (C1). In contrast, we observed a relatively lower frequency of errors related to empty pixel sets (C2) or calculation failures (C3).
Overall, the low accuracy suggests that existing LLMs are not yet capable of reliably producing code for Earth Observation tasks, likely due to the under-representation of this specific domain in pre-training corpora. Consequently, \datasetname serves as a practically relevant, out-of-domain benchmark for future research aimed at overcoming these limitations.

\begin{figure}[th!]
  \centering
  \includegraphics[width=0.95\linewidth]{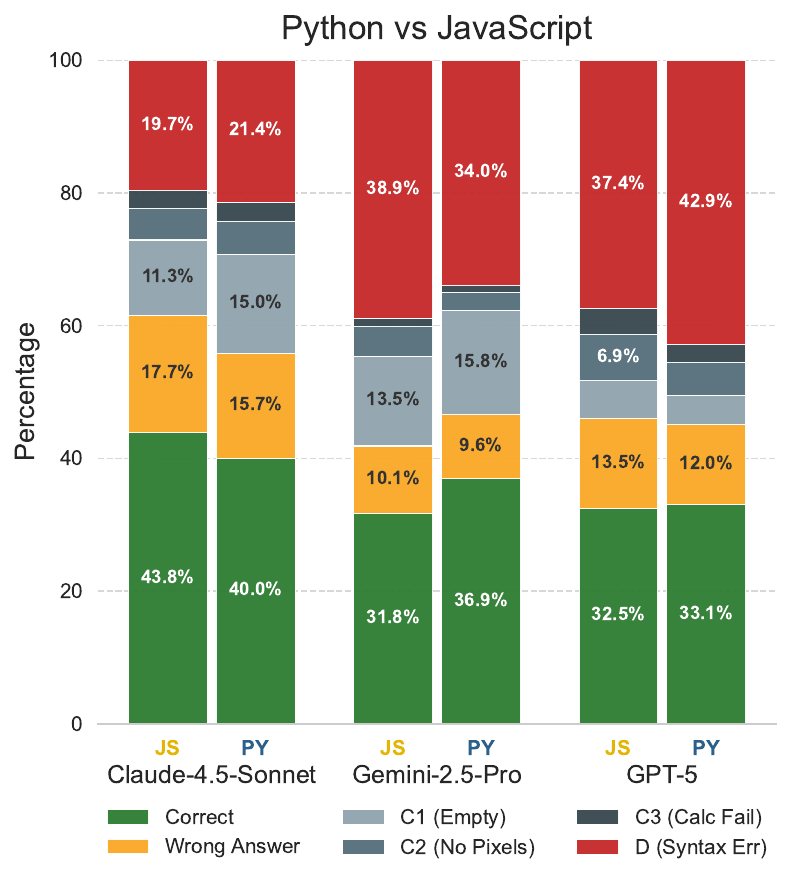}
  \caption{Performance in Python against JavaScript.}
  \label{fig:python_vs_js}
\end{figure}

\begin{figure*}[th!]
\centering
\includegraphics[width=0.95\linewidth]{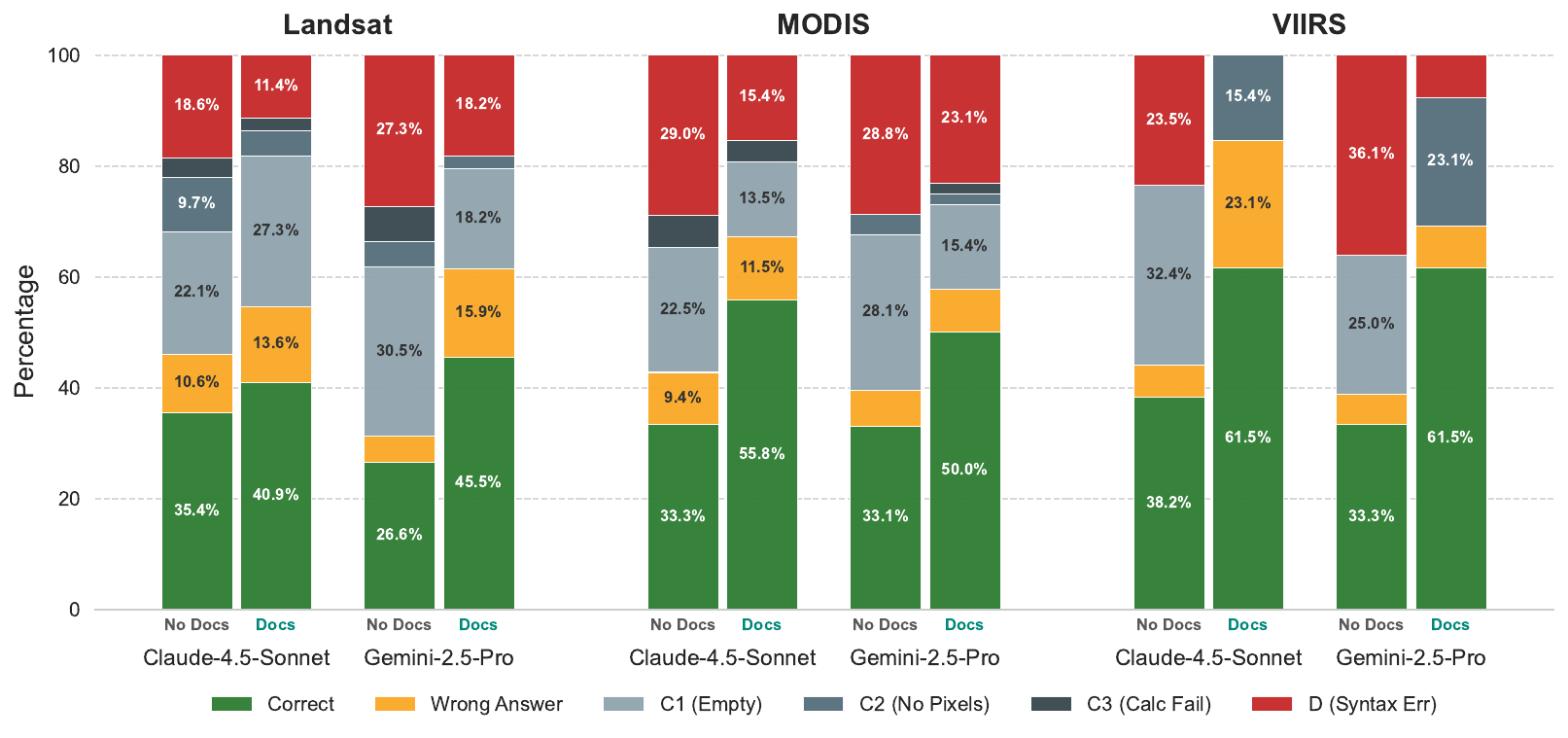}
\caption{Impact of providing ground-truth Collection IDs (documentation) on performance, stratified by instrument. 
}
\label{fig:documentation_impact}
\end{figure*}

\subsection{Python versus JavaScript}

Since Google Earth Engine (GEE) also offers a JavaScript API for rapid prototyping and interactive analysis, we benchmark whether model performance is language-dependent. As illustrated in Figure~\ref{fig:python_vs_js}, models exhibit distinct sensitivities to the choice of programming language. Claude-4.5-Sonnet achieves a slight performance gain, rising from 40.0\% in Python to 43.8\% in JavaScript. Conversely, performance for Gemini-2.5-Pro and GPT-5 degrades by 5.1\% and 0.6\%, respectively, when using JavaScript. Overall, LLM agents do show a comparable performance regarding API languages.

Investigating the cause, we find that the logic-based wrong answer rate remains consistent across languages. Instead, the divergence stems from execution stability. For Claude, the improvement is driven by a better ability to identify valid image collections in JavaScript (i.e., reduce in type D error). In contrast, the performance drop in Gemini-2.5-Pro models is primarily caused by an increase in syntax errors, indicating a relative unfamiliarity with the GEE JavaScript API compared to its Python counterpart.

\begin{figure*}[th]
\centering
\includegraphics[width=0.95\linewidth]{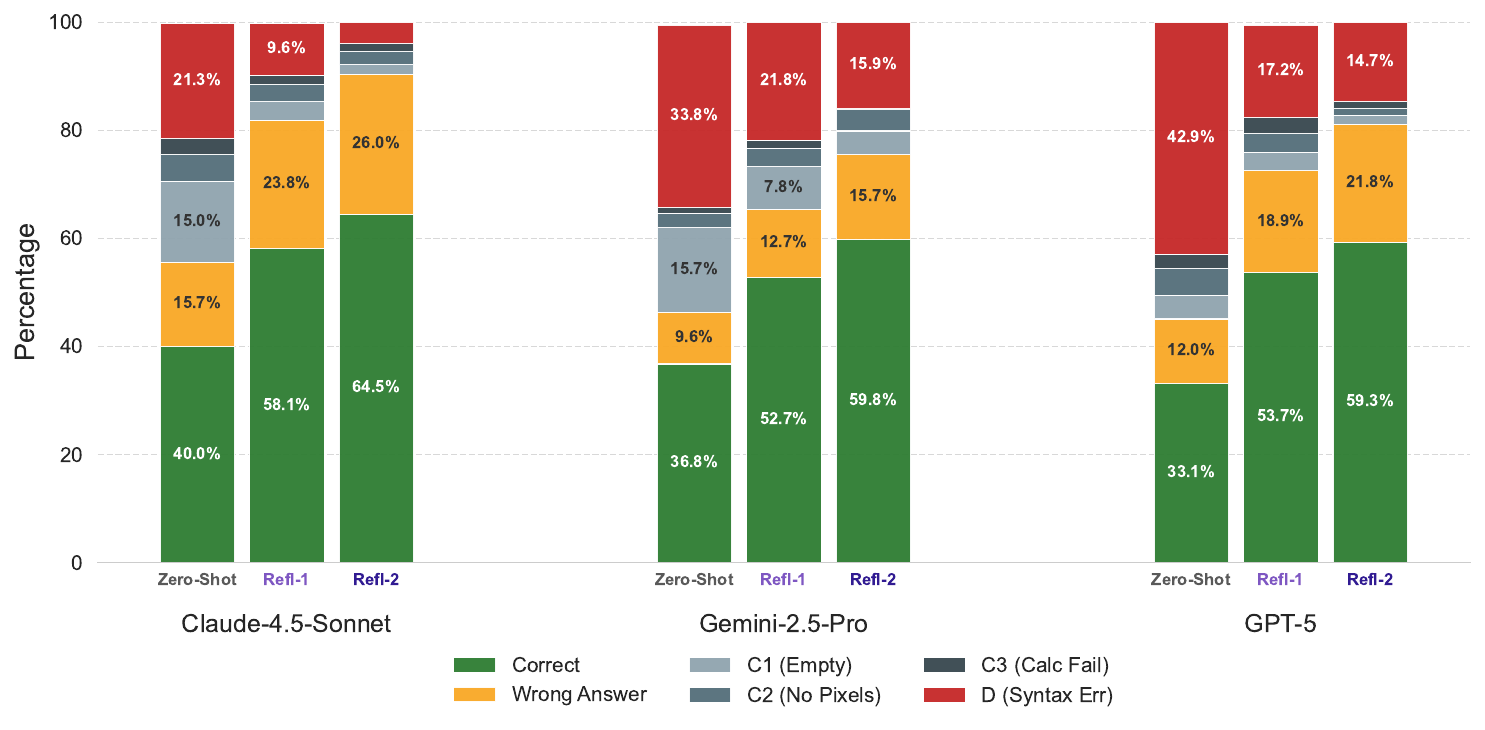}
\caption{Reflexion improves performance. We observe a consistent increase in correctness across all agents. This gain is primarily driven by the effective mitigation of code and syntax errors (Category D).
}
\label{fig:reflexion_impact}
\end{figure*}

\subsection{Does Providing Collection ID Reduce Hallucinations?}

Given the high prevalence of hallucinated dataset names observed in Table~\ref{tab:id_hallucinations}, we investigate whether providing specific GEE Collection IDs and metadata in the prompt mitigates these errors. We focus our analysis on questions resolvable via Landsat, MODIS, and VIIRS modalities.
To this end, we adopt a two-round dialogue approach. In the first round, the agent is asked to select the appropriate modality ("Landsat", "MODIS", "VIIRS", or "others") for the query. Based on this selection, the associated documentation, including the precise Collection ID and band names, is injected into the prompt for the second round, where the model generates the executable code. If "others" is selected, no additional information is provided.

The stratified results are presented in Figure~\ref{fig:documentation_impact}. 
The provision of documentation yields a clear reduction in Code/Syntax Errors (Category D) across nearly all configurations. For instance, in the MODIS modality, Claude-4.5-Sonnet reduces its syntax error rate from 29.0\% to 15.4\%. This effect is most pronounced in the VIIRS modality, where Claude-4.5-Sonnet effectively eliminates syntax errors (23.5\% to 0\%). Consequently, VIIRS accuracy improves significantly from 38.2\% to 61.5\%.

However, we observe that a reduction in syntax errors does not always translate directly to correctness. As execution success rates rise, other error categories become more visible. In the VIIRS case, while the code runs successfully, the Wrong Answer rate increases from 8\% to 23.1\% for Claude-4.5-Sonnet. In the case of Gemini-2.5-Pro on VIIRS, the errors change from category C2 (gray) and D (red) to C3 (darker gray).
These findings suggest that while documentation enables the model to successfully access the data, it does not necessarily ensure the full correctness of the coding.

In summary, while documentation successfully lowers the barrier to data access (resolving hallucinations), it exposes secondary challenges in logical data processing. Once the syntax hurdles are cleared, models often produce valid code that fails to retrieve data due to incorrect date filtering or cloud masking logic, shifting the failure mode from an immediate crash (Category D) to an empty result (Category C1).

\subsection{Does Reflection Help?}

To assess the potential for improvement beyond the zero-shot baseline, we leverage the \textit{reflexion} framework~\cite{shinn2023reflexion}. In our experimental setup, the reflection mechanism is triggered specifically when the model fails to produce a valid execution result (i.e., encountering syntax errors, timeouts, or empty returns), rather than on logical correctness alone. During the reflection phase, the agent is presented with the original query, its previous code attempt, and the resulting execution traceback or error log. It is then instructed to debug and regenerate the script. We allow for a maximum of two rounds of such reflection.

As illustrated in Figure~\ref{fig:reflexion_impact}, the reflexion scheme show substantial performance gains across all evaluated models. {Claude-4.5-Sonnet} demonstrates the highest overall efficacy, improving its correctness from {40.0\%} in the zero-shot setting to {64.5\%} after two rounds of reflection. GPT-5 also exhibits a dramatic recovery, jumping from a baseline of 33.1\% to 59.3\%, a 79 percentage point increase.

The breakdown of error categories reveals the mechanics of this improvement. The feedback loop is particularly effective at mitigating Code/Syntax Errors (Category D). We also observe a notable reduction in Empty Images (C1) errors, suggesting that the error logs effectively guide the models to correct invalid date filters or choose the correct modalities that have the data for the specific time.

However, it is worth noting that as syntax and execution errors decrease, the rate of wrong answers increases simultaneously and proportionally (e.g., rising from 15.7\% to 26.0\% for Claude-4.5-Sonnet). This indicates that while self-reflection allows the code to run to completion, it does not guarantee that the code logic is correct. Nevertheless, the ability to convert fatal errors into executable code significantly boosts the overall ceiling for autonomous Earth Observation analysis.

%% file: tables/bl_gee.tex
\begin{table*}[th]
\centering
\caption{Performance Comparison of Code LLMs on Earth Engine Tasks. The table reports the distribution of outcomes across distinct categories. \textbf{Execution Success} indicates the code ran without errors (Correct vs. Wrong Answer). \textbf{Failed Execution/Logic} indicates specific failure modes: Empty Collection or Images (C1), No Valid Pixels (C2), Calculation Failure (C3), or Syntax Error (D). 
}
\vspace{-1.7mm}
\label{table:gee_bl}
\footnotesize
\begin{tabular}{l cc cccc}
\toprule
& \multicolumn{2}{c}{Execution Success (\%)} & \multicolumn{4}{c}{Failed Execution / Logic (\%)} \\
\cmidrule(lr){2-3} \cmidrule(lr){4-7}
Model 
& \textbf{Correct} 
& Wrong Ans.
& Empty (C1) 
& No Pixels (C2) 
& Calc Fail (C3) 
& Syntax (D) \\
\midrule
Gemini-2.5-Pro & \cellcolor{ForestGreen!50}36.9 & 9.6 & 15.8 & 2.7 & 1.0 & 34.0 \\
Gemini-2.5-Flash & \cellcolor{ForestGreen!10}10.6 & 5.9 & 15.0 & 4.4 & 3.2 & 60.8 \\
\midrule
Claude-4.5-Sonnet & \cellcolor{ForestGreen!60}\textbf{40.0} & 15.7 & 15.0 & 4.9 & 2.9 & 21.4 \\
Claude-4.5-Haiku & \cellcolor{ForestGreen!20}16.7 & 12.8 & 14.0 & 4.9 & 4.9 & 46.7 \\
\midrule
GPT-5 & \cellcolor{ForestGreen!40}33.1 & 12.0 & 4.4 & 4.9 & 2.7 & 42.9 \\
\midrule
DeepSeek-Reasoner & \cellcolor{ForestGreen!30}28.5 & 13.0 & 15.0 & 5.2 & 4.2 & 34.2 \\
DeepSeek-Chat & \cellcolor{ForestGreen!25}27.0 & 13.0 & 17.0 & 7.4 & 4.9 & 30.7 \\
\midrule
Kimi-k2 & \cellcolor{ForestGreen!20}17.0 & 8.4 & 23.1 & 3.7 & 1.2 & 46.7 \\
\bottomrule
\end{tabular}
\end{table*}

%% file: tables/collection_hallucination.tex
\begin{table}[t] 
    \centering
    \caption{Comparison of hallucinated vs. correct GEE Collection IDs. Models often struggle with specific naming conventions (e.g., version numbers, tiers, or provider prefixes).}
    \label{tab:id_hallucinations}
    \vspace{-1.7mm}
    \footnotesize 
    \setlength{\tabcolsep}{3pt} 
    \renewcommand{\arraystretch}{1.2} 

    \begin{tabularx}{\linewidth}{@{} l X @{}} 
        \toprule
        \textbf{Domain} & \textbf{Collection ID ( \xmark~Hallucinated / \cmark~Correct )} \\
        \midrule
        
        Landsat & \xmark~\texttt{LANDSAT/LC08/C02/SR} \\
                & \cmark~\texttt{LANDSAT/LC08/C02/T1\_L2} \\ 
        \addlinespace[4pt] 
        
        VIIRS & \xmark~\texttt{NOAA/VIIRS/001/VNP14IMGTDL\_NRT} \\
              & \cmark~\texttt{NASA/LANCE/SNPP\_VIIRS/C2} \\
        \addlinespace[4pt]

        TRMM & \xmark~\texttt{NASA/TRMM/3B43V7} \\
             & \cmark~\texttt{TRMM/3B43V7} \\
        \addlinespace[4pt]

        MODIS & \xmark~\texttt{MODIS/051/MCD19A2\_GRANULAR\_AOD} \\
              & \cmark~\texttt{MODIS/061/MCD19A2\_GRANULES} \\
        \addlinespace[4pt]

        JRC & \xmark~\texttt{JRC/GHSL/P2023A/GHS\_POP} \\
            & \cmark~\texttt{JRC/GHSL/P2023A/GHS\_SMOD\_POP} \\
        \addlinespace[4pt]


        \bottomrule
    \end{tabularx}
\end{table}

%% file: sec/main_6_conclusion.tex
\section{Conclusion}
We introduce \datasetname to investigate how reliable are AI agents in solving Earth observation questions.
When asked to produce evidence in the form of Google Earth Engine (GEE) code, the best zero-shot accuracy is 40.0\% primarily because of models' inability to correctly navigate the many data sources. 
Our detailed analysis identifies clear pathways for improvement. We show that the primary bottleneck is not a lack of coding capability, but a lack of domain-specific grounding. By injecting precise documentation regarding Collection IDs, we significantly reduce hallucinations, though this unmasks deeper challenges in downstream logical reasoning. Furthermore, we find that employing iterative self-correction (reflexion) can recover a substantial portion of these failures, boosting performance to 64.5\%. By curating \datasetname, we hope to expose the gap between general-purpose reasoning and domain-specific execution, paving the way for future research into agents that can reason scientifically about the Earth.

%% file: sec/main_7_appendice.tex

\section{More Dataset Description}
\label{app:dataset_description}

\input{tables/sec3_table_joint}
Our dataset also comprises a large variety ($\sim$ 17) of satellite sensors.
MODIS (Moderate Resolution Imaging Spectroradiometer) observations are most numerous due to its daily temporal resolution and complementary morning and afternoon observations from MODIS Terra and MODIS Aqua satellites. The second highest is Landsat (Land Remote-Sensing Satellite), which has provided historical coverage dating to 1972, making it valuable for decade-long comparisons and analyses, though its 16-day revisit time limits temporal resolution. VIIRS (Visible Infrared Imaging Radiometer Suite), launched in 2012, offers daily observations with specialized capabilities for nighttime light intensity measurements. 
Note that in Earth observation, we refer to the instrument as a sensor and its data products as products; for Google Earth Engine (GEE), these are organized as imagery collections.

Other important sensors, though less frequently mentioned in the posts, include: TRMM (Tropical Rainfall Measuring Mission) for precipitation monitoring; GRACE (Gravity Recovery and Climate Experiment) for gravity field measurements; TOMS (Total Ozone Mapping Spectrometer) for atmospheric ozone monitoring; SMAP (Soil Moisture Active Passive) for global soil moisture mapping; GLDAS (Global Land Data Assimilation System) for land surface modeling. These sensors, while appearing less frequently in our dataset, play crucial roles in long-term Earth observation and environmental monitoring.

\section{Introduction to Google Earth Engine}
\label{app:gee_intro}

Google Earth Engine (GEE)~\citep{gorelick2017google} is a cloud-based platform that enables users to perform geospatial analysis at a planetary scale using Google's computational infrastructure. It houses over 90 petabytes of analysis-ready satellite imagery and more than 1,000 curated geospatial datasets spanning 50+ years of historical data, including imagery from satellites such as Landsat, MODIS, and Sentinel, as well as climate and weather datasets, geophysical data, terrain information, and land cover data. Researchers harness this technology for various Earth Observation and applications such as forest mapping~\citep{chen2017mangrove}, drought monitoring~\citep{sazib2018leveraging}, crop yield estimation~\citep{jaafar2021gymee}, land use and land cover~\citep{nasiri2022land}, evapotranspiration~\citep{senay2022mapping}, shoreline analysis~\citep{santra2024quantifying}, and water detection~\citep{yue2023fully}, etc.

\textbf{Terminology.} 
A \textbf{sensor} refers to a device that detects and measures physical properties (like reflectance, temperature, etc.), such as optical cameras, radar, and spectrometers mounted on satellites or aircraft.
A \textbf{product} is a processed dataset derived from sensor data, typically preprocessed for calibration, quality control, and transformation into specific variables. In GEE specifically, an \textbf{imagery collection} is a set of related images grouped together for analysis.

\textbf{API Usage.} GEE provides a JavaScript and Python API that enables users to access and filter the extensive data catalog, apply algorithms for image processing and analysis, perform geospatial computations across multiple processors in parallel. In this paper, the AI agents generate the Python code and execute it. The GEE API script calls the GEE server for the computation. The results, mostly the final statistics, are sent back to the local agents for further deduction and answering.

\section{Error Analysis}

\paragraph{Error Taxonomy (Gemini-2.5-Pro, Zero-Shot Baseline).}
Table~\ref{tab:error_baseline} shows the breakdown of failure modes under the zero-shot baseline. This reveals that the dominant failure modes are knowledge-related, not code-related:
wrong dataset IDs (14.2\%), temporal/spatial mismatch (15.7\%), and wrong band names (9.6\%)
together account for $\sim$40\% of all failures, while only 10\% are pure syntax errors.

\begin{table*}[th]
\centering
\begin{tabular}{lrr}
\toprule
\textbf{Error Type} & \textbf{Count} & Percentage(\textbf{\%}) \\
\midrule
Correct                                        & 150 & 36.8\% \\
Wrong Dataset (hallucinated asset ID)          &  58 & 14.2\% \\
Temporal/Spatial Mismatch (empty collection)   &  64 & 15.7\% \\
Wrong Band (non-existent band name)            &  39 &  9.6\% \\
Syntax/Runtime Error                           &  41 & 10.0\% \\
Wrong Answer (correct execution, wrong reasoning) & 39 & 9.6\% \\
Masking/Preprocessing (no valid pixels)        &  11 &  2.7\% \\
Numerical Failure (NaN/None result)            &   4 &  1.0\% \\
\bottomrule
\end{tabular}
\caption{Error taxonomy for Gemini-2.5-Pro under the zero-shot baseline.}
\label{tab:error_baseline}
\end{table*}

After applying iterative re-planning (\textsc{Oscar}~\cite{wang2024oscar}, 3 rounds), the error profile (Table~\ref{tab:error_replanning} shifts significantly. Re-planning substantially reduces knowledge-grounding errors (wrong dataset, wrong band,
temporal/spatial mismatch) and execution errors. However, the proportion of wrong answers
more than doubles from 9.6\% to 21.3\%: as more questions successfully execute, the remaining
errors shift from pipeline failures to reasoning errors where the code runs correctly but
produces an incorrect conclusion (\emph{e.g.}, wrong threshold, incorrect comparison logic).
This suggests that after resolving surface-level failures, the next frontier is improving the
agent's domain reasoning and scientific judgment.

\begin{table*}[th]
\centering
\begin{tabular}{lrrr}
\toprule
\textbf{Error Type} & \textbf{Baseline} & \textbf{After Re-planning} & \textbf{ $\Delta$} \\
\midrule
Correct                    & 36.8\% & 55.9\% & $+$19.1\% \\
Wrong Dataset              & 14.2\% & 10.3\% & $-$3.9\%  \\
Wrong Band                 &  9.6\% &  1.5\% & $-$8.1\%  \\
Temporal/Spatial Mismatch  & 15.7\% &  2.9\% & $-$12.8\% \\
Syntax/Runtime             & 10.0\% &  2.5\% & $-$7.5\%  \\
Wrong Answer               &  9.6\% & 21.3\% & $+$11.7\% \\
\bottomrule
\end{tabular}
\caption{Error profile shift after iterative re-planning (\textsc{Oscar}, 3 rounds).}
\label{tab:error_replanning}
\end{table*}

This process-level analysis provides actionable diagnostic insight:
(1) \emph{retrieval-augmented tool grounding}, where agents query a structured catalog at
inference time rather than relying on memorized schemas, and
(2) \emph{improved domain reasoning} through better prompting strategies or domain-specific
fine-tuning to address the growing proportion of reasoning errors.

%% file: sec/main_7_annotation.tex
\newpage
\section{Reviewer instructions for \datasetname}
\label{app:evaluation_criteria}
Below, we detail the instructions given to the reviewers.
\subsection{Goal}
Given a question and an article about earth science, your task is to provide an answer.

\subsection{Evaluation Questions}
\begin{enumerate}
    \item \textbf{What is the answer to this question?}\\
    Please answer (A) Yes, (B) No, or (C) I don't know, or data is not conclusive.
    \item \textbf{Is the answer to the question being supported by the text from the article?}\\
    Please copy and paste the relevant texts that you use to derive your answer from the article.
    \begin{itemize}
        \item \textit{Strongly Supported}: The article explicitly states the answer or provides clear evidence.
        \item \textit{Moderately Supported}: The article implies the answer, but requires some inference.
        \item \textit{Not Supported}: The article contradicts the answer or provides no relevant information.
    \end{itemize}
    \item \textbf{Is the answer to the question being supported by the image from the article?}\\
    If yes, please explain how the image supports the answer to the question.
    \begin{itemize}
        \item \textit{Strongly Supported}: The article explicitly states the answer or provides clear evidence.
        \item \textit{Moderately Supported}: The article implies the answer, but requires some inference.
        \item \textit{Not Supported}: The article contradicts the answer or provides no relevant information.
    \end{itemize}
    \item \textbf{Do you need to use Google Maps to check location information?}\\
    If yes, please explain why using Google Maps is required.
    Please answer (A) Yes, (B) No.
\end{enumerate}

Below are the three examples provided to the reviewers. 

\begin{tcolorbox}[colback=gray!10, colframe=gray!40, title=Example 1]
\textbf{Question:} Does the Tuolumne River Basin have more snow on April 1, 2017 than on April 1, 2015?\\
\textbf{URL:} \href{https://earthobservatory.nasa.gov/images/90073/sierra-snowpack-bigger-than-last-four-years-combined}{link}
    
\textbf{Q1: What is the answer to this question?}\\
    You should answer \textbf{(A) Yes}
    
\textbf{Q2: Is the answer to the question being supported by the text from the article?}\\
    You should answer \textbf{Strongly Supported}\\
    You should copy the text \textit{"New NASA data show that snowpack in Tuolumne River Basin—a major source of water for San Francisco and California's Central Valley—is currently greater than that of the four previous years combined."} and paste it to the spreadsheet.
    
\textbf{Q3: Is the answer to the question being supported by the image from the article?}\\
    You should answer \textbf{Strongly Supported}\\
    You should explain the reason: \textit{"The image shows greater snow water equivalent in April 1, 2017, compared to April 1, 2015"}
    
\textbf{Q4: Do you need to use Google Maps to check location information?}\\
    You should answer \textbf{No}.
    
\textbf{Q5: Other comments}\\
    You don't have to write anything.
\end{tcolorbox}


\begin{tcolorbox}[colback=gray!10, colframe=gray!40, 
fontupper=\small,
title=Example 2,
]
\textbf{Question:} Does Cape Lookout National Seashore show lower turbidity in the region centered at (34.659539, -76.464976) than the region centered at (34.607982, -76.338262) on February 18, 2016?\\
\textbf{URL:} \href{https://earthobservatory.nasa.gov/images/87627/on-the-lookout}{link}

\textbf{Q1: What is the answer to this question?}\\
    You should answer \textbf{(B) No}
    
\textbf{Q2: Is the answer to the question being supported by the text from the article?}\\
    You should answer \textbf{Not Supported}\\
    You should write \textit{"The text does not mention that"}.
    
\textbf{Q3: Is the answer to the question being supported by the image from the article?}\\
    You should answer \textbf{Strongly Supported}\\
    You should explain the reason: \textit{"The image shows that (34.659539, -76.464976) had less turbidity than another region"}
    
\textbf{Q4: Do you need to use Google Maps to check location information?}\\
    You should answer \textbf{Yes}\\
    You should explain the reason: \textit{"Neither the image nor the text shows the two geolocations."}
    
\textbf{Q5: Other comments}\\
    You don't have to write anything.
\end{tcolorbox}

\begin{tcolorbox}[colback=gray!10, colframe=gray!40, title=Example 3]
\textbf{Question:} Does Cape Lookout National Seashore show lower turbidity in the region centered at (34.659539, -76.464976) than the region centered at (34.607982, -76.338262) on February 18, 2017?\\
\textbf{URL:} \href{https://earthobservatory.nasa.gov/images/87627/on-the-lookout}{link}

\textbf{Q1: What is the answer to this question?}\\
    You should answer \textbf{(C) I don't know, or data is not conclusive}
    
\textbf{Q2: Is the answer to the question being supported by the text from the article?}\\
    You should answer \textbf{Not Supported}\\
    You should write \textit{"The text talks about events in 2016, not 2017"}.
    
\textbf{Q3: Is the answer to the question being supported by the image from the article?}\\
    You should answer \textbf{Not Supported}\\
    You should explain the reason: \textit{"The time period is incorrect."}
    
\textbf{Q4: Do you need to use Google Maps to check location information?}\\
    You should answer \textbf{Yes}\\
    You should explain the reason: \textit{"Neither the image nor the text shows the two geolocations."}
    
\textbf{Q5: Other comments}\\
    You can write \textit{"I think the question is wrong. Please take a look."}
\end{tcolorbox}